\documentclass[10pt,twocolumn,letterpaper]{article}

\usepackage{iccv}
\usepackage{times}
\usepackage{epsfig}
\usepackage{graphicx}
\usepackage{amsmath}
\usepackage{amssymb}
\usepackage{setspace}

\if CLASSOPTIONcompsoc
\usepackage[caption=false, font=normalsize, labelfont=sf, textfont=sf]{subfig}
\else
\usepackage[caption=false, font=footnotesize]{subfig}
\fi

\usepackage[pagebackref=true,breaklinks=true,letterpaper=true,colorlinks,bookmarks=false]{hyperref}

\newcommand{\1}[1]{{\bf{\color{red}#1}}}
\newcommand{\2}[1]{{\bf{\color{blue}#1}}}

\usepackage{pifont}

\newcommand*{\affaddr}[1]{#1} % No op here. Customize it for different styles.

\newcommand*{\email}[1]{\texttt{#1}}

\iccvfinalcopy % *** Uncomment this line for the final submission

\def\iccvPaperID{4765} % *** Enter the ICCV Paper ID here

\ificcvfinal\fi

\begin{document}

%%%%%%%%% TITLE
\title{Dynamic Curriculum Learning for Imbalanced Data Classification}

\author{
Yiru Wang \thanks{Equal contribution.}\,\,,\, Weihao Gan$^{~*}$,\, Jie Yang,\, Wei Wu,\, Junjie Yan\\
\affaddr{SenseTime Group Limited}\\
% \affaddr{\affmark[2]Beihang University}\\
\email{\tt\small \{wangyiru,ganweihao,yangjie,wuwei,yanjunjie\}@sensetime.com}\\%
}

\maketitle
\ificcvfinal\thispagestyle{empty}\fi

%%%%%%%%% ABSTRACT

%%%%%%%%% ABSTRACT
\begin{abstract}
	Human attribute analysis is a challenging task in the field of computer vision. One of the significant difficulties is brought from largely imbalance-distributed data. Conventional techniques such as re-sampling and cost-sensitive learning require prior-knowledge to train the system. To address this problem, we propose a unified framework called Dynamic Curriculum Learning (DCL) to adaptively adjust the sampling strategy and loss weight in each batch, which results in better ability of generalization and discrimination. Inspired by curriculum learning, DCL consists of two-level curriculum schedulers: (1) sampling scheduler which manages the data distribution not only from imbalance to balance but also from easy to hard; (2) loss scheduler which controls the learning importance between classification and metric learning loss. With these two schedulers, we achieve state-of-the-art performance on the widely used face attribute dataset CelebA and pedestrian attribute dataset RAP.
\end{abstract}

%%%%%%%%% BODY TEXT
\section{Introduction}
Human attribute analysis, including facial characteristics and clothing categories, has facilitated the society in various aspects, such as tracking and identification. However, different from the general image classification problem like ImageNet challenge \cite{krizhevsky2012imagenet}, human attribute analysis naturally involves largely imbalanced data distribution. For example, when collecting the face data of attribute `Bald', most of them would be labeled as `No Bald' and its imbalanced ratio to the `Bald' class would be relatively high. Training the classification model with equal importance for samples in different classes may result in a bias to the majority class of the data and poor accuracy for the minority class. Therefore, it is of great importance to handle the imbalanced data learning problem, especially in human attribute analysis.

% imbalanced ratio to (?)

Impressive results have been achieved for the general imbalanced data learning in the past years. One intuitive motivation is resampling \cite{chawla2002smote, drummond2003c4, han2005borderline, he2008learning, maciejewski2011local,he2008adasyn}, which either oversamples the minority class data or downsamples the majority class data, to balance the data distribution. However, oversampling could easily cause overfitting problem due to repeatedly visiting duplicated minority samples, while downsampling may discards much useful information in the majority samples. Another kind of approach called cost-sensitive learning is also exploited to handle the imbalanced data learning problem, which  directly imposes heavier cost on the misclassified minority class \cite{ tang2009svms, ting2000comparative,  zadrozny2003cost, zhou2006training} . However, it is difficult to determine the exact cost for different samples in various distributions. Hand et al. \cite{sl} proposed  a batch-wise method  that selects part of the majority samples and increases the weight of minority samples to match a pre-defined target distribution. Besides  the standard cross entropy classification loss, Dong et al. \cite{crl1, crl} proposed to add  another class rectification loss (CRL) to avoid the dominant effect of majority classes. 
A specific metric is proposed for imbalanced datasets by above methods. For the general classification problem, \textbf{class-biased accuracy} is defined as the number of correctly predicted samples divided by the number of the whole test data. While for imbalanced data classification, \textbf{class-balanced accuracy} is defined as the average of the accuracy in each class for evaluation.

Our proposed Dynamic Curriculum Learning (DCL) method is motivated by the following two considerations. (1) Sampling is an acceptable strategy for the problem, but keeping targeting at a balanced distribution in the whole process would hurt the generalization ability, particularly for a largely imbalanced task. For example, in the early stage of learning with balanced target distribution, the system discards lots of majority samples and emphasizes too much on the minority samples, tending to learn the valid representation of the minority class but the bad/unstable representation of the majority one. However, what we expect is to make the system first learn the appropriate general representations for both of the classes on the target attributes and then classify the samples into correct labels, which results in a favorable balance between the class bias accuracy and class balanced accuracy. (2) It is reasonable to combine cross entropy loss (CE) and metric learning loss (ML) since the appropriate feature representation could be helpful for classification. However, we think those two components contribute to different emphasis. Treating them equally in the training process cannot fully utilize the discriminative power of deep CNN. Specifically, CE pays more attention to the classification task by assigning specific labels, while ML focuses more on learning a soft feature embedding to separate different samples in feature space without assigning labels. Similarly to the previous point, we expect the system first to learn the appropriate feature representation and then classify the samples into the correct labels.

In the spirit of the curriculum learning \cite{curr}, we propose Dynamic Curriculum Learning (DCL) framework for imbalanced data learning. Specifically, we design two-level curriculum schedulers: (1) sampling scheduler: it aims to find the most meaningful samples in one batch to train the model dynamically from imbalanced to balanced and from easy to hard; (2) loss scheduler: it controls the learning weights between classification loss and metric learning loss. These two components can be defined by the scheduler function, which reflects the model learning status. To summarize our contributions:
\begin{itemize}
	\setlength{\itemsep}{0pt}
\setlength{\parsep}{0pt}
\setlength{\parskip}{0pt}
	\item For the first time, we introduce the curriculum learning idea into  imbalanced data learning problem. Based on the designed scheduler function, two curriculum schedulers are proposed for dynamic sampling operation and loss backward propagation.
	\item The proposed DCL framework is a unified representation, which can generalize to several existing state-of-the-art methods with corresponding setups.
	\item We achieve the new state-of-the-art performance on the commonly used face attribute dataset CelebA \cite{liu2015faceattributes} and pedestrian attribute dataset RAP \cite{li2016richly}.
\end{itemize}

%------------------------------------------------------------------------
\section{Related Work}
\textbf{Imbalanced data learning.} There are several groups of methods trying to address the imbalanced learning problem in literature. (1) Data-level: considering the imbalanced distribution of the data, one intuitive way to do is resampling the data \cite{ chawla2002smote, drummond2003c4, han2005borderline, he2008learning, maciejewski2011local, oquab2014learning,he2013imbalanced,estabrooks2004multiple} into a balanced distribution, which could oversample the minority class data and downsample the majority class data. One advanced sampling method called SMOTE \cite{chawla2002smote,chawla2004special} augments artificial examples created by interpolating neighboring data points. Some extensions of this technique were proposed \cite{han2005borderline, maciejewski2011local}. However, oversampling can easily cause overfitting problem due to repeatedly visiting duplicated minority samples. While downsampling usually discards many useful information in majority samples. (2) Algorithm-level: cost-sensitive learning aims to avoid above issues by directly imposing a heavier cost on misclassifying the minority class \cite{ tang2009svms, ting2000comparative,  zadrozny2003cost, zhou2006training,yang2009margin,thai2010cost}. However, how to determine the cost representation in different problem settings or environments is still an open question. Besides of the cost-sensitive learning, another option is to change the decision threshold during testing, which is called threshold-adjustment technique \cite{chen2006decision, yu2016odoc, zhou2006training}.
(3) Hybrid: this is an approach that combines multiple techniques from one or both abovementioned categories. Widely used example is ensembling idea. EasyEnsemble and BalanceCascade are methods that train a committee of classifiers on undersampled subsets \cite{liu2009exploratory}. SMOTEBoost, on the other hand, is a combination of boosting and SMOTE oversampling \cite{chawla2003smoteboost}. Some methods like \cite{napierala2010learning,khoshgoftaar2011comparing,seiffert2014empirical,van2009knowledge,zhang2015active,pan2013graph} also pays attention to the noisy samples in the imbalanced dataset.

\textbf{Deep imbalanced learning.} Recently, several deep methods have been proposed for imbalanced data learning \cite{  crl1, crl, sl, lmle, CLMLE, jeatrakul2010classification, khan2018cost, khoshgoftaar2010supervised, shen2015deepcontour,  zhou2006training,sarafianos2018deep,cui2019class}. One major direction is to integrate the sampling idea and cost-learning into an efficient end-to-end deep learning framework. Jeatrakul et al. \cite{jeatrakul2010classification} treated the Complementary Neural Network as an under-sampling technique, and combined it with SMOTE-based over-sampling to rebalance the data. Zhou et al. \cite{zhou2006training} studied data resampling for training cost-sensitive neural networks. In \cite{ khan2018cost,cui2019class}, the cost-sensitive deep features and the cost parameter are jointly optimized. Oquab et al. \cite{oquab2014learning} resampled the number of foreground and background image patches for learning a convolutional neural network (CNN) for object classification.  Hand et al. \cite{sl} proposed a selective learning(SL) method to manage the sample distribution in one batch to a target distribution and assign larger weight for minority classes for backward propagation. Another recent direction of the problem involves the metric learning into the system.  Dong et al. \cite{crl1, crl} proposed a class rectification loss (CRL) regularising algorithm to avoid the dominant effect of majority classes by discovering sparsely sampled boundaries of minority classes. More recently, LMLE/CLMLE \cite{lmle, CLMLE} are proposed to preserve the local class structures by enforcing large margins between intra-class and inter-class clusters.

\textbf{Curriculum learning.} The idea of curriculum learning was originally proposed in \cite{curr}, it demonstrates that the strategy of learning from easy to hard significantly improves the generalization of the deep model.  Up to now, works been done via curriculum learning mainly focus on visual category discovery \cite{lee2011learning,sarafianos2018curriculum}, object tracking \cite{supancic2013self}, semi-/weakly-supervised learning \cite{gong2016multi, guo2018curriculumnet, jiang2015self, pentina2015curriculum}, etc. \cite{pentina2015curriculum} proposed an approach that processes multiple tasks in a sequence with sharing between subsequent tasks instead of solving all tasks jointly by finding the best order of tasks to be learned. Very few works approach the imbalanced learning. Guo et al. \cite{guo2018curriculumnet} developed a principled learning strategy by leveraging curriculum learning in a weakly supervised framework, with the goal of effectively learning from imbalanced data.

\section{Method}

We propose a Dynamic Curriculum Learning (DCL) framework for imbalanced data classification problem, consisting of two-level \textbf{curriculum schedulers}. 
The first one is a sampling scheduler of which the key idea is to find the most significant samples in one batch to train the model dynamically making data distribution from imbalanced to balanced and from easy to hard. This scheduler determines the sampling strategy for the proposed Dynamic Selective Learning (DSL) loss function.
The second one is the loss scheduler, which controls the learning importance between two losses: the DSL loss and the metric learning loss (triplet loss). Therefore, in the early stage of the training process, the system focuses more on the soft feature space embedding, while later on, it pays more attention to the task of classification.

\subsection{Scheduler Function Design} \label{SFD}
Most of the traditional curriculum learning methods manually define different training strategies. While in our proposed DCL framework for imbalanced data learning, we formulate the key idea of curriculum scheduling with different groups of functions, as we called \textbf{Scheduler Function}. We show the semantic interpretation for those functions.

\begin{figure}

	\centering
	\includegraphics[width=.33\textwidth]{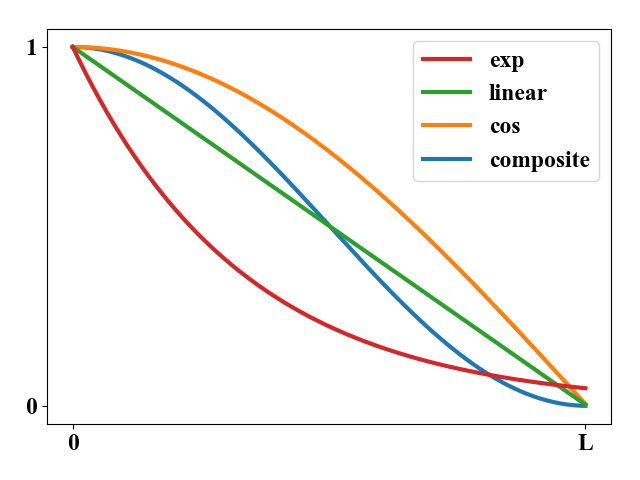}
	\caption{Four types of designed scheduler functions.
	}
	\label{img:gx}
\end{figure}

The scheduler function $SF(l)$ is a function which returns value monotonically decreasing from 1 to 0 with the input variable $l$, which represents the current training epoch. It reflects the model learning status and measures the curriculum learning speed. We explore several function classes as following (illustrated in Figure \ref{img:gx}):
% , $L$ refers to expected total training epochs
\begin{spacing}{0.5}
	\begin{itemize}
		\item {Convex function: indicating the learning speed from slow to fast. For example:}
		      \begin{equation}
			      SF_{cos}(l) = cos (\frac{l}{L} * \frac{\pi}{2})
			      \label{cos_g}
		      \end{equation}

		\item {Linear function: indicating the constant learning speed. For example:}
		      \begin{equation}
			      SF_{linear}(l) = 1- \frac{l}{L}
			      \label{linear_g}
		      \end{equation}

		\item {Concave function: indicating the learning speed from fast to slow. For example:}
		      \begin{equation}
			      SF_{exp}(l) = \lambda^{l}
			      \label{exp_g}
		      \end{equation}

		\item {Composite function:  indicating the learning speed from slow to fast and then slow again. For example:}
		      \begin{equation}
			      SF_{composite}(l) = \frac{1}{2} cos(\frac{l}{L}\pi)+ \frac{1}{2}
			      \label{comp_g}
		      \end{equation}
	\end{itemize}
	%%行间距变为single-space
\end{spacing}
\noindent
where $L$ refers to expected total training epochs and $\lambda$ is an independent hyperparameter that in the range of $(0,1)$.

%---------------------------------------------------------------------------------

%---------------------------------------------------------------------------------

Different classes of $SF(l)$ represent different curriculum learning styles. Based on the above-introduced scheduler functions, we propose Dynamic Curriculum Learning framework for imbalanced data classification.

\subsection{Sampling Scheduler}

Sampling is one of the most commonly used techniques to deal with imbalanced data learning.  In this section, we introduce the proposed Dynamic Selective Learning (DSL) component, which is based on our sampling scheduler. The sampling scheduler dynamically adapts the target distribution in a batch from imbalanced to balanced during the training process.

Explicitly, for each attribute, we define $j^{th}$ element of the data distribution $D$ as the number of $j^{th}$ class samples divided by the number of minority samples (the least one). Sorting them in ascending order, then we have:
\begin{equation}
			D = 1: \frac{\#C_{1}}{\#C_{min}}:  \frac{\#C_{2}}{\#C_{min}} : ... :  \frac{\#C_{K-1}}{\#C_{min}}
\end{equation}
\noindent
where $K$ is the number of classes and $\#C_{i}$ is the number of samples in class $i$. Each attribute has its training distribution $D_{train}$, which is a global statistic.

Sampling scheduler determines the target data distribution of the attributes in each batch. Initially, the target distribution of one attribute $D_{target}(0)$ in a batch is set to $D_{train}$, which is imbalanced distributed. During the training process, it gradually transfers to a balanced distribution with the following function (each element is powered by $g(l)$):
\begin{spacing}{0.7}
	\begin{equation}
		D_{target}(l)={D_{train}}^{g(l)}
		\label{eq:power}
	\end{equation}
\end{spacing}
\noindent
where $l$ refers to current training epoch and $g(l)$ is the sampling scheduler function, which can be any choice in Section \ref{SFD}. According to target distribution $D_{target}(l)$ , the majority class samples are dynamically selected and the minority class samples are re-weighted in different epochs to confirm different target distributions in one batch. Therefore, the DSL loss is defined as:
\begin{spacing}{0.5}
	\begin{equation}
		\small
		\mathcal{L}_\text{DSL} = -\frac{1}{N}\sum_{j=1}^{M}\sum_{i=1}^{N_{j}}w_{j}*\textup{log}\left ( p(y_{i,j}=\bar{y}_{i,j} | \textbf{x}_{i,j}) \right )
		\label{eq:dsl}
	\end{equation}

	\begin{equation}
		\small
		w_{j} =
		\begin{cases}
			\frac{D_{target,j}(l)}{D_{current,j}} & \mbox{if }  \frac{D_{target,j}(l)}{D_{current,j}} \geq 1 \\
			0/1                                   & \mbox{if } \frac{D_{target,j}(l)}{D_{current,j}} < 1
		\end{cases}
		\label{eq:w}
	\end{equation}
\end{spacing} 
\noindent
where $N$ is batch size, $N_{j}$ is the number of samples of $j^{th}$ class in current batch, $M$ is number of classes, $\bar{y}_{i,j}$ is the ground truth label. $w_{j}$ is the cost weight for class $j$. $D_{target,j}(l)$ is the $j^{th}$ class target distribution in current epoch $l$. $D_{current,j}$ is the $j^{th}$ class distribution in current batch before sampling. If $\frac{D_{target,j}(l)}{D_{current,j}} < 1$, we sample $\frac{D_{target,j}(l)}{D_{current,j}}$ percentage of $j^{th}$ class data with original weight 1 and the remainings with 0. If not, then $j^{th}$ class is a minority class and a larger weight is assigned to the  samples.

With different sampling scheduler functions (four types in the previous section), the batch target distribution changes from the training set biased distribution to balanced distribution. At the beginning epoch, $g(0)=1$, the target distribution $D$ equals to the train set distribution; in other words, the real-world distribution. At the final epoch, $g(l)$ is close to 0, so all the element in target distribution $D$ is close to 1 (power of 0). In other words, it is a balanced distribution.

The learning rate is usually set conforming to a decay function. At the early stage of the training process, with a large learning rate and biased distribution, the curriculum scheduler manages the model to learn more on whole training data. Usually, the system learns lots of easy samples in this stage. Going further with the training process, the target distribution is gradually getting balanced. With the selected majority samples and re-weighted minority samples, the system focuses more on the harder situation.

\subsection{Metric Learning with Easy Anchors}

Besides of the loss function$\mathcal{L}_\text{DSL}$, we also involve a metric learning loss to learn a better feature embedding for imbalance data classification.

A typical selection of the metric learning loss is triplet loss, which was introduced by CRL\cite{crl} with hard mining. Define the samples with high prediction score on the wrong class as hard samples. Then we build triplet pairs from the anchors and some hard positive and negative samples. The loss function  in CRL is defined as following:
%---------------------------------------------------------------------------------
\begin{spacing}{0.8}
	\begin{equation}
		\small\mathcal{L}_\text{crl}  = \frac{\sum_{T} 	\max \Big(0, \,\, m_j +{d}(\textbf{x}_{all,j},\textbf{x}_{+,j}) - {d}(\textbf{x}_{all,j},
		\textbf{x}_{-,j}) \Big)}{|T|}
		\label{eq:crl}
	\end{equation}
\end{spacing}
%---------------------------------------------------------------------------------
\noindent
where $m_j$ refers to the margin of class $j$ in triplet loss and $d(\cdot)$ denotes the feature distance between two samples. In current batch, $\textbf{x}_{all,j}$ represents all the samples in class j, $\textbf{x}_{+,j}$ and $\textbf{x}_{-,j}$ represents positive samples and negative samples respectively. T refers to the number of triplet pairs. In CRL\cite{crl}, all the minority class samples are selected as anchors.

\begin{figure}
	\centering

	\includegraphics[width=.42\textwidth]{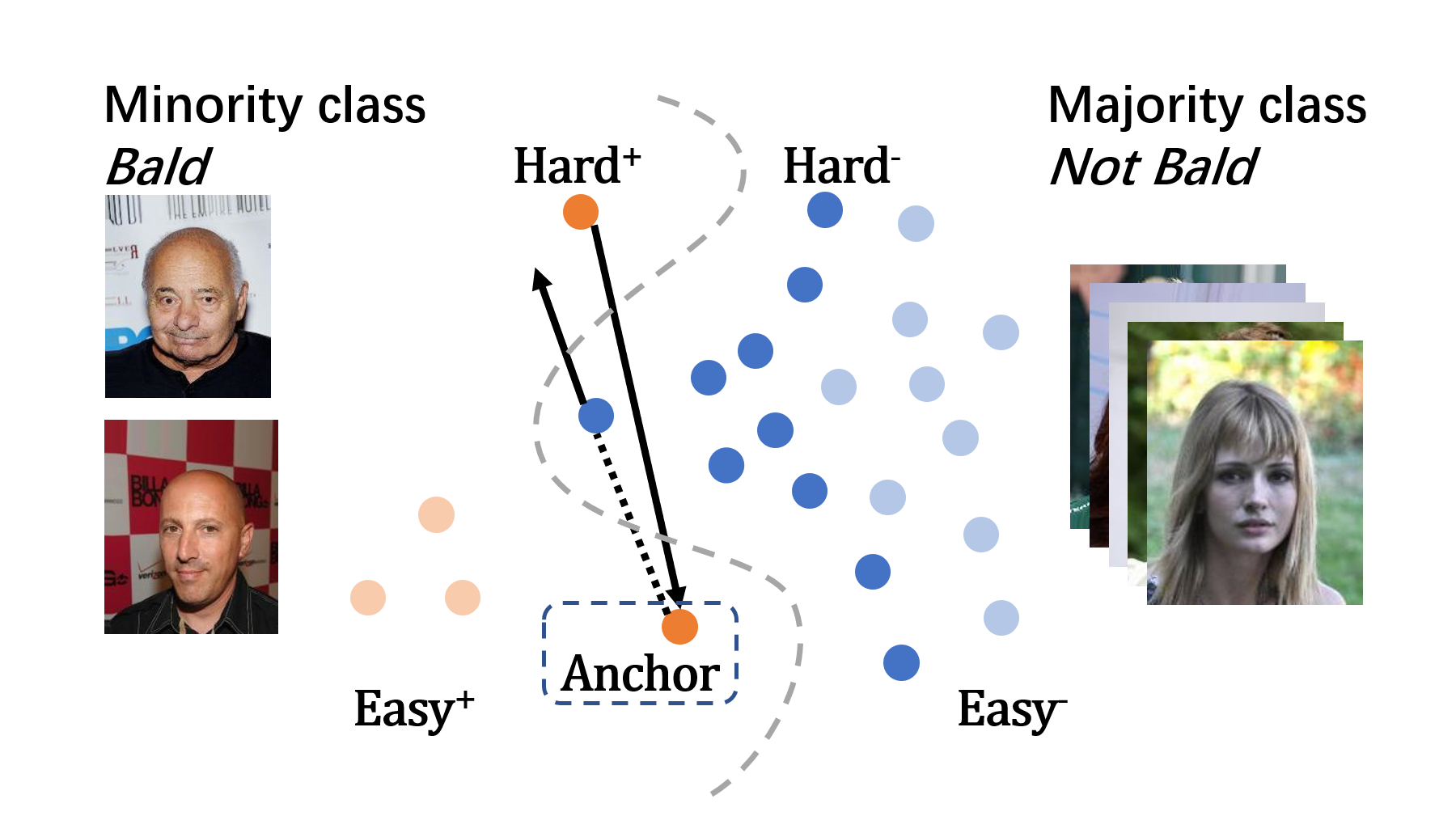}
	\caption{This figure visualizes a case of Triplet Loss in CRL\cite{crl} that hard positive sample is chosen as the anchor.  Assuming minority class as the positive class, the triplet pair shown in the figure is trying to push both the positive sample and the negative sample across the border, which is pushing the positive sample closer to the negative side. It can cause the features of positive samples to be more chaotic. }
	\label{hard_anchor}

	\vspace{2ex}

	\centering
	\includegraphics[width=.42\textwidth]{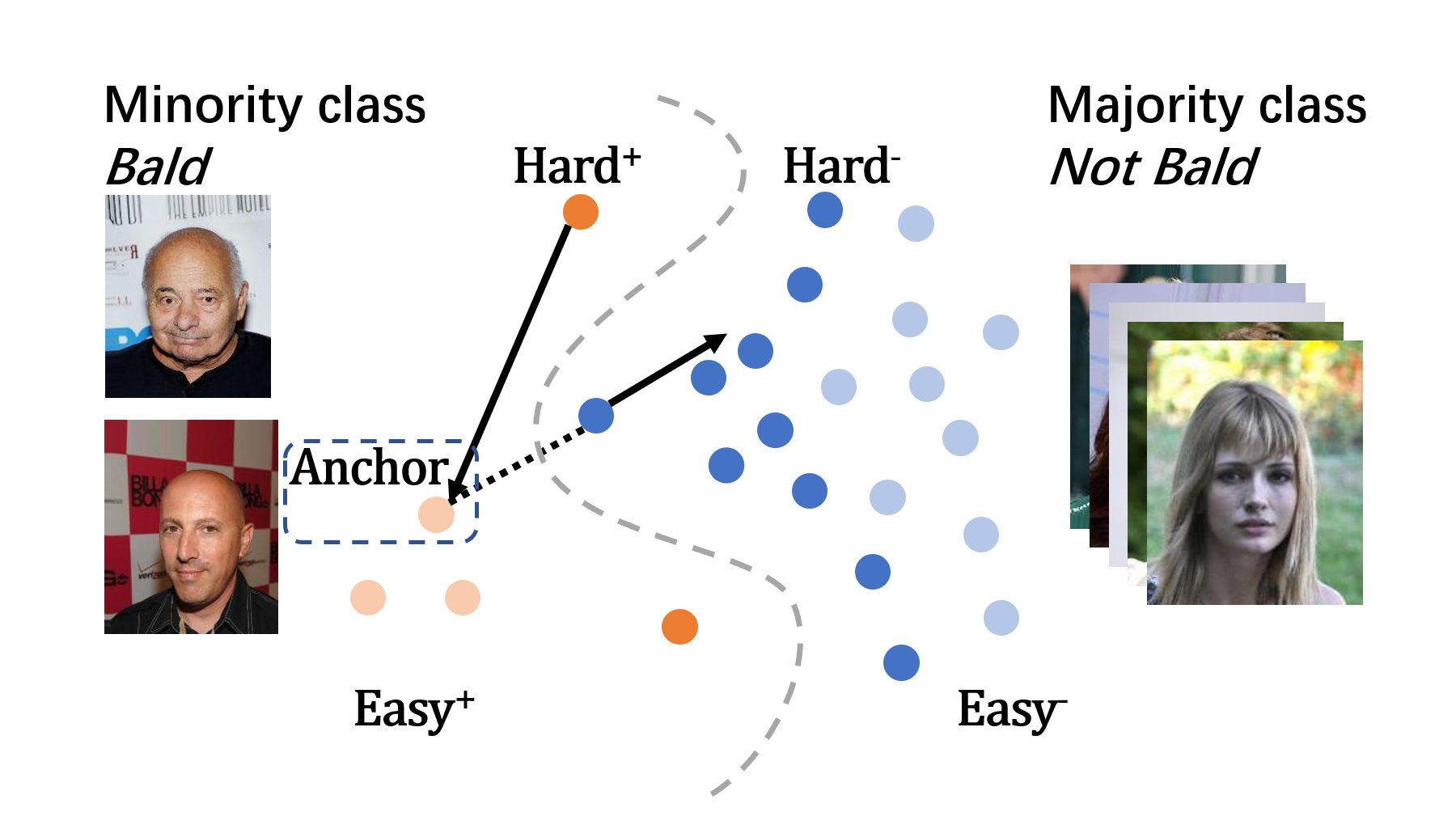}
	\caption{This figure visualizes a case of our proposed Triplet Loss with only easy positive samples as the anchor. Since easy positive samples' features can be grouped easily, the hard positive sample can be pulled closer to all the easy positive samples. Our proposed method can avoid the situation
		in Figure \ref{hard_anchor}. }
	\label{easy_anchor}

\end{figure}

We define \textbf{easy sample} as the correctly predicted sample. Choosing all the minority samples as anchors is not stable for model to learn, since it may cause problems such as pulling easy positive samples to the negative side. Examples are illustrated in Figure \ref{hard_anchor}.

We propose a method to improve the sampling operation of Triplet loss with Easy Anchors $\mathcal{L}_\text{TEA} $, defined as follow:
%---------------------------------------------------------------------------------
\begin{equation}
	\small
	%\scriptsize
	%\footnotesize
	\mathcal{L}_\text{TEA}  = \frac{\sum_{T} 	\max \Big(0, \,\, m_j +{d}(\textbf{x}_{easy,j},\textbf{x}_{+,j}) - {d}(\textbf{x}_{easy,j},
	\textbf{x}_{-,j}) \Big)}{|T|}
	%{|X_\text{min}| \times n_\text{attr} \times K^2}
	%
	%\sum_{T}
	%\max(0, m + \mbox{dist}(\bm{x}_a^i,\bm{x}_p^i) - \mbox{dist}(\bm{x}_a^{i}, \bm{x}_n^i)) 
	%
	\label{eq:tea}
\end{equation}
%---------------------------------------------------------------------------------
\noindent
where $\textbf{x}_{easy,j}$ refers to easy minority samples in class $j$, others are similar to  equation \ref{eq:crl}. Easy anchors are defined as high-confident correctly predicted minority samples. The number of hard positives,  hard negatives and easy anchors to be selected is determined by the hyper-parameter $k$.

With $\mathcal{L}_\text{TEA}$ loss, only easy samples in minority class are chosen as anchors, which pulls the hard positive samples closer and pushes hard negative samples further. As illustrated in Figure \ref{easy_anchor}.  Different from CRL choosing all minority samples as anchors to make rectification on feature space, our proposed method selects easy anchors based on the result of the classifier and pull all the samples to well-classified side. Also, we adopt the hard sample mining for those selected easy anchors to build the triplet loss.

\subsection{Loss Scheduler}

To train the model better, we analyze the different characteristics of the two proposed losses. Generally speaking, triplet loss targets at learning a soft feature embedding to separate different samples in feature space without assigning labels, while cross entropy loss aims to classify the samples by assigning specific labels.

Particularly for imbalanced data learning, what we want is that the system first learns an appropriate feature representation then benefits the classification. Therefore, in order to fully utilize these two properties, we design a loss curriculum scheduler $f(l)$  to manage these two losses.

Even though we can choose any one of the schedule functions in Section \ref{SFD}, we use the composite function (Equation \ref{comp_g}) as an example here. The model learns with the following scheduler:
\begin{spacing}{0.8}
	%---------------------------------------------------------------------------------
	\begin{equation}
		\mathcal{L}_\text{DCL}  = \mathcal{L}_\text{DSL} + f(l) * \mathcal{L}_\text{TEA}
		\label{eq:triplet_weight}
	\end{equation}
	%---------------------------------------------------------------------------------

	%---------------------------------------------------------------------------------
	\begin{equation}
		\small
		%\scriptsize
		%\footnotesize
		f(l) =
		\begin{cases}
			\frac{1}{2} cos(\frac{l}{L}\pi)+ \frac{1}{2}+\epsilon & \mbox{if }
			l < pL                                                                       \\
			\epsilon                                              & \mbox{if } l \geq pL
		\end{cases}
		\label{eq:triplet_weight}
	\end{equation}
	%---------------------------------------------------------------------------------
\end{spacing}
\noindent
where $l$ refers to current training epoch, $L$ refers to expected total training epochs. Small modifications including a hyperparameter $p$ ranging in $[0,1]$, which is defined as advanced self-learning point. Moreover, $\epsilon$ is the self-learning ratio. The reason why we have a non-zero $\epsilon$ here is that even though in self-learning stage, the model still needs to maintain the feature structure learned from in the previous stages.

In the early stage of training, a large weight is initialized to the triplet loss $\mathcal{L}_\text{TEA}$ for learning soft feature embedding and decreases through time in respect to the scheduler function. In the later stage, the scheduler assigns a small impact on $\mathcal{L}_\text{TEA}$ and system emphasizes more on the Dynamic Selective Loss $ \mathcal{L}_\text{DSL} $ to learn the classification. Finally, when it reaches the self-learning point, no `teacher' curriculum scheduler is needed. The model automatically finetunes the parameters until convergence.

%---------------------------------------------------------------------------------
%\begin{figure}
%
%	\centering
%	\includegraphics[width=.33\textwidth]{fx.png}
%	\caption{Network Loss Scheduler}
%	\label{img:fx}
%\end{figure}
%---------------------------------------------------------------------------------

\subsection{Generalization of DCL Framework}

To handle the imbalanced data learning problem, we propose the Dynamic Curriculum Learning framework. Revisiting the overall system, DCL consists of two-level curriculum schedulers. One is for sampling $g(l)$, and another is for loss learning $f(l)$. We can find that several state-of-the-art imbalanced learning methods can be generalized from the framework with different setups for the schedulers. The correspondings are listed in Table \ref{tb:general}. Selective Learning \cite{sl} does not contain metric learning and only uses a fixed target distribution. CRL-I\cite{crl1} does not contain a re-weight or re-sample operation and only uses a fixed weight for metric learning.

\begin{table}[]
	\centering
	\resizebox{0.48\textwidth}{!}{
		\begin{tabular}{|c|c|c|}
			\hline
			Method                      & g(x)               & f(x)           \\ \hline
			Cross Entropy               & 1                  & 0              \\ \hline
			Selective Learning\cite{sl} & 0/1                & 0              \\ \hline
			CRL-I\cite{crl1}            & 1                  & $\epsilon$     \\ \hline
			DCL(Ours)                   & Sampling scheduler & Loss scheduler \\ \hline
		\end{tabular}

	}
	\vspace{3pt}
	\caption{Generalization of proposed Dynamic Curriculum Learning method to other non-clustering imbalanced learning methods with corresponding setups. }
	\label{tb:general}
\end{table}

\begin{table*}[t]
	\caption{Class-balanced Mean Accuracy (mA) for each class (\%) and class imbalance level (majority class rate-50\%) of each of the attributes on CelebA dataset. The 1st/2nd best results are highlighted in red/blue.
	}
	\centering
	\resizebox{0.95\textwidth}{!}{
		\large
		\begin{tabular}{c|c|c|c|c|c|c|c|c|c|c|c|c|c|c|c|c|c|c|c|c|c}
			\hline
			                                    & \rotatebox[origin=lB]{90}{Attractive}
			                                    & \rotatebox[origin=lB]{90}{Mouth Open}
			                                    & \rotatebox[origin=lB]{90}{Smiling}
			                                    & \rotatebox[origin=lB]{90}{Wear Lipstick}
			                                    & \rotatebox[origin=lB]{90}{High Cheekbones}
			                                    & \rotatebox[origin=lB]{90}{Male}
			                                    & \rotatebox[origin=lB]{90}{Heavy Makeup}
			                                    & \rotatebox[origin=lB]{90}{Wavy Hair}
			                                    & \rotatebox[origin=lB]{90}{Oval Face}
			                                    & \rotatebox[origin=lB]{90}{Pointy Nose}
			                                    & \rotatebox[origin=lB]{90}{Arched Eyebrows}
			                                    & \rotatebox[origin=lB]{90}{Black Hair}
			                                    & \rotatebox[origin=lB]{90}{Big Lips}
			                                    & \rotatebox[origin=lB]{90}{Big Nose}
			                                    & \rotatebox[origin=lB]{90}{Young}
			                                    & \rotatebox[origin=lB]{90}{Straight Hair}
			                                    & \rotatebox[origin=lB]{90}{Brown Hair}
			                                    & \rotatebox[origin=lB]{90}{Bags Under Eyes}
			                                    & \rotatebox[origin=lB]{90}{Wear Earrings}
			                                    & \rotatebox[origin=lB]{90}{No Beard}
			                                    & \rotatebox[origin=lB]{90}{Bangs}                                                                                                                                                                                                                                 \\
			\hline \hline
			Imbalance level                     & 1                                            & 2                  & 2      & 3      & 5      & 8      & 11     & 18     & 22     & 22     & 23             & 26     & 26     & 27     & 28     & 29     & 30     & 30     & 31     & 33     & 35                 \\ \hline
			DeepID2(CE)~\cite{deepid2}          & 78                                           & 89                 & 89     & 92     & 84     & 94     & 88     & 73     & 63     & 66     & 77             & 83     & 62     & 73     & 76     & 65     & 79     & 74     & 75     & 88     & 91                 \\  \hline
			Over-Sampling~\cite{drummond2003c4}
			                                    & 77                                           & 89                 & 90     & 92     & 84     & 95     & 87     & 70     & 63     & 67
			                                    & 79                                           & 84                 & 61     & 73     & 75     & 66     & 82     & 73     & 76     & 88     & 90                                                                                                                   \\

			Down-Sampling~\cite{drummond2003c4}
			                                    & 78                                           & 87                 & 90     & 91     & 80     & 90     & 89     & 70     & 58     & 63
			                                    & 70                                           & 80                 & 61     & 76     & 80     & 61     & 76     & 71     & 70     & 88     & 88                                                                                                                   \\

			Cost-Sensitive~\cite{he2008learning}
			                                    & 78                                           & 89                 & 90     & 91     & 85     & 93     & 89     & 75     & 64     & 65
			                                    & 78                                           & 85                 & 61     & 74     & 75     & 67     & 84     & 74     & 76     & 88     & 90                                                                                                                   \\

			Selective-Learning~\cite{sl}        & 81                                           & 91                 & 92     & 93     & 86     & 97     & 90     & 78     & 66     & 70     & 79             & 87     & 66     & 77     & 83     & 72     & 84     & 79     & 80     & 93     & 94                 \\

			CRL-I~\cite{crl1}                   & 83                                           & 95                 & 93     & 94     & 89     & 96     & 84     & 79     & 66     & \2{73} & 80             & \2{90} & \2{68} & \2{80} & 84     & 73     & 86     & 80     & 83     & 94     & 95                 \\

			LMLE~\cite{lmle}                    & \2{88}                                       & \2{96}             & \1{99} & \1{99} & \2{92} & \1{99} & \1{98} & \2{83} & 68     & 72     & 79             & 92     & 60     & \2{80} & \2{87} & 73     & \2{87} & 73     & 83     & \2{96} & \2{98}             \\

			CLMLE~\cite{CLMLE}                  & \1{90}                                       & \1{97}             & \1{99} & \2{98} & \1{94} & \1{99} & \1{98} & \1{87} & \1{72} & \1{78} & \1{86}         & \1{95} & 66     & \1{85} & \1{90} & \1{80} & \1{89} & \1{82} & \1{86} & \1{98} & \1{99}             \\ \hline

			DCL~(ours)                          & 83                                           & 93                 & 93     & 95     & 88     & 98     & 92     & 81     & \2{70} & \2{73} & \2{82}         & 89     & \1{69} & \2{80} & 86     & \2{76} & 86     & \1{82} & \2{85} & 95     & 96                 \\

			\hline \hline
			                                    & \rotatebox[origin=lB]{90}{Blond Hair}
			                                    & \rotatebox[origin=lB]{90}{Bushy Eyebrows}
			                                    & \rotatebox[origin=lB]{90}{Wear Necklace}
			                                    & \rotatebox[origin=lB]{90}{Narrow Eyes}
			                                    & \rotatebox[origin=lB]{90}{5 o'clock Shadow}
			                                    & \rotatebox[origin=lB]{90}{Receding Hairline}
			                                    & \rotatebox[origin=lB]{90}{Wear Necktie}
			                                    & \rotatebox[origin=lB]{90}{Eyeglasses}
			                                    & \rotatebox[origin=lB]{90}{Rosy Cheeks}
			                                    & \rotatebox[origin=lB]{90}{Goatee}
			                                    & \rotatebox[origin=lB]{90}{Chubby}
			                                    & \rotatebox[origin=lB]{90}{Sideburns}
			                                    & \rotatebox[origin=lB]{90}{Blurry}
			                                    & \rotatebox[origin=lB]{90}{Wear Hat}
			                                    & \rotatebox[origin=lB]{90}{Double Chin}
			                                    & \rotatebox[origin=lB]{90}{Pale Skin}
			                                    & \rotatebox[origin=lB]{90}{Gray Hair}
			                                    & \rotatebox[origin=lB]{90}{Mustache}
			                                    & \rotatebox[origin=lB]{90}{Bald}
			                                    &
			                                    & \rotatebox[origin=lB]{90}{\textbf{Average}}                                                                                                                                                                                                                      \\
			\hline \hline
			Imbalance level                     & 35                                           & 36                 & 38     & 38     & 39     & 42     & 43     & 44     & 44     & 44     & 44             & 44     & 45     & 45     & 45     & 46     & 46     & 46     & 48     &        &                    \\ \hline
			DeepID2(CE)~\cite{deepid2}          & 90                                           & 78                 & 70     & 64     & 85     & 81     & 83     & 92     & 86     & 90     & 81             & 89     & 74     & 90     & 83     & 81     & 90     & 88     & 93     &        & \textbf{81.17}     \\  \hline

			Over-Sampling~\cite{drummond2003c4} & 90                                           & 80                 & 71     & 65     & 85     & 82     & 79     & 91     & 90     & 89     & 83             & 90     & 76     & 89     & 84     & 82     & 90     & 90     & 92     &        & \textbf{81.48}     \\

			Down-Sampling~\cite{drummond2003c4}
			                                    & 85                                           & 75                 & 66     & 61     & 82     & 79     & 80     & 85     & 82     & 85
			                                    & 78                                           & 80                 & 68     & 90     & 80     & 78     & 88     & 60     & 79     &        & \textbf{77.45}                                                                                                       \\

			Cost-Sensitive~\cite{he2008learning}
			                                    & 89                                           & 79                 & 71     & 65     & 84     & 81     & 82     & 91     & \1{92} & 86
			                                    & 82                                           & 90                 & 76     & 90     & 84     & 80     & 90     & 88     & 93     &        & \textbf{81.60}                                                                                                       \\\

			Selective-Learning~\cite{sl}        & 93                                           & 85                 & 73     & \2{74} & 89     & \2{87} & 92     & 97     & 90     & 94     & \2{87}         & \2{94} & \2{86} & 96     & \2{89} & 92     & 94     & 92     & 95     &        & \textbf{85.93}     \\

			CRL-I~\cite{crl1}                   & 95                                           & 84                 & \2{74} & 72     & 90     & \2{87} & 88     & 96     & 88     & 96     & \2{87}         & 92     & 85     & 98     & \2{89} & 92     & 95     & \2{94} & \2{97} &        & \textbf{86.60}     \\

			LMLE~\cite{lmle}                    & \1{99}                                       & 82                 & 59     & 59     & 82     & 76     & 90     & 98     & 78     & 95     & 79             & 88     & 59     & \1{99} & 74     & 80     & 91     & 73     & 90     &        & \textbf{83.83}     \\

			CLMLE~\cite{CLMLE}                  & \1{99}                                       & \1{88}             & 69     & 71     & \2{91} & 82     & \1{96} & \1{99} & 86     & \1{98} & 85             & \2{94} & 72     & \1{99} & 87     & \2{94} & \2{96} & 82     & 95     &        & \2{\textbf{88.78}} \\ \hline

			DCL~(ours)                          & 95                                           & \2{87}             & \1{76} & \1{79} & \1{93} & \1{90} & \2{95} & \1{99} & \1{92} & \2{97} & \1{93}         & \1{97} & \1{93} & \1{99} & \1{94} & \1{96} & \1{99} & \1{97} & \1{99}
			                                    &                                              & \1{\textbf{89.05}}                                                                                                                                                                                                \\ \hline
		\end{tabular}
	}
	\label{tb1}
	\vspace{-1em}
\end{table*}

\section{Experiments}

\subsection{Datasets}

\noindent
\textbf{CelebA} \cite{liu2015faceattributes} is a human facial attribute dataset with annotations of 40 binary
classifications. CelebA is an imbalanced dataset, specifically on some attributes, where the sample imbalance level (majority class rate-50$\%$) could be up to 48. The dataset contains 202,599 images from 10,177 different people.

\noindent
\textbf{RAP} \cite{li2016richly} is a richly annotated dataset for pedestrian attribute recognition in real surveillance scenario. It contains 41,585 images from 26 indoor cameras, with 72 different attributes. RAP is a highly imbalanced dataset with the imbalance ratio (minority samples to majority samples)  up to 1:1800.

%%add
\noindent
\textbf{CIFAR-100} \cite{krizhevsky2009cifar} is a natural image classification dataset with $32\times32$ pixels. It contains 50,000 images for training and 10,000 images for testing. It is a balanced dataset with 100 classes. Each class holds the same number of images.

%%data? 

\subsection{Evaluation Metric}

For CelebA dataset and RAP dataset, following the standard profile, we apply the class-balanced accuracy (binary classification) on every single task, and then compute the mean accuracy of all tasks as the overall metric. It can be formulated as following:
\begin{spacing}{0.8}
	\begin{equation}
		mA_i=\frac{1}{2}(\frac{TP_i}{P_i}+\frac{TN_i}{N_i})
	\end{equation}
	\begin{equation}
		mA=\frac{\Sigma_{i=1}^{|C|} mA_i}{|C|}
		\label{ma}
	\end{equation}
\end{spacing}
\noindent
where $mA_i$ indicates the class-balanced mean accuracy of the $i$-th task, with $TP_i$ and $P_i$ indicating the count of predicted true positive samples and positive samples in the ground truth for the $i$-th task while $TN_i$ and $N_i$ refers to the opposite. ${|C|}$ is the number of tasks.

For CIFAR-100 dataset, since each class holds the same number of instances, class-balanced accuracy equals to class-biased accuracy.

\subsection{Experiments on CelebA Face Dataset}
\subsubsection{Implementation Details}

\textbf{Network Architecture}
We use DeepID2\cite{deepid2} as the backbone for experiments on CelebA for a fair comparison. DeepID2\cite{deepid2} is a CNN of 4 convolution layers. All the experiments listed on table \ref{tb1}  set DeepID2\cite{deepid2} as backbone. The baseline is trained with a simple Cross-Entropy loss. Since CelebA is a multi-task dataset, we set an independent 64D feature layer and a final output layer for each task branch. For each branch, it considers its own current and target distribution and generates single attribute loss (Equation \ref{eq:triplet_weight}). Then we sum them up for backpropagation in a joint-learn fashion. 
%All the experiments listed on table \ref{tb1} without extra declearation set DeepID2\cite{deepid2} as backbone.
% DeepID2\cite{deepid2} is an obsolete network with limited capacity, so we slightly modified its structure for further comparison. In the last row of table \ref{tb1}, we changed the pooling layers to convolution layers  in order to increase the capacity of the network.

\noindent
\textbf{Hyper-Parameter Settings} We train DCL at learning rate of 0.003, batch size at 512, training epoch at 300 and weight decay at 0.0005. Horizontal Flip is applied during training. Specifically, we set sampling scheduler to convex function in Equation \ref{cos_g}, loss scheduler to composite function in Equation \ref{eq:triplet_weight} with advanced self-learning point $p$  to 0.3, and $k$ in $\mathcal{L}_{TEA}$ (Equation \ref{eq:tea}) to 25. The margin is set to 0.2.

\noindent
\textbf{Time Performance} We train all the models with TITAN XP GPU. Compared to the baseline DeepID2 which takes 20 hours to train, DCL training framework spends 20.5 hours to converge (only 0.5 hour more on sampling and loss calculation) under the same 300 epochs.

\subsubsection{Overall Performance}

We compared our proposed method DCL with DeepID2 \cite{deepid2}, Over-Sampling and Down-Sampling in \cite{drummond2003c4}, Cost-Sensitive \cite{he2008learning}, Selective Learning (SL) \cite{sl}, CRL\cite{crl1}, LMLE\cite{lmle} and CLMLE\cite{CLMLE}.

Table \ref{tb1} shows the overall results on CelebA. The baseline of our evaluation is the general face classification framework DeepID2\cite{deepid2} with standard cross entropy loss, where we achieve around 8$\%$ performance improvement. Compared to the recent advanced method, our method outperforms 3.12$\%$ to Selective Learning\cite{sl}, 2.45$\%$ to CRL-I\cite{crl1}, 5.22$\%$ to LMLE\cite{lmle} and 0.27$\%$ to CLMLE\cite{CLMLE}, respectively. Specifically, LMLE/CLMLE methods are sample-clustering based methods. However, one sample is usually bundled with multiple different attributes. It is challenging  to handle all the aspects of different attributes in constructing quintuplet (four-samples). In our proposed DCL method, it treats different attributes individually  based on their own distributions and the triplet loss is also defined in attribute-level so that it can be easily expanded to multiple attributes learning problem. Besides, our method is computational efficient with minimal extra time cost compared to the cross-entropy loss. In LMLE/CLMLE, a computational expensive data pre-processing (including clustering and quintuplet construction) is required for each round of deep model learning. To create a quintuplet for each data sample, four cluster- and class-level searches are needed. 

\subsubsection{Effect of Data Imbalance Level}

%---------------------------------------------------------------------------------
\begin{figure}

	\centering
	\includegraphics[width=.43\textwidth]{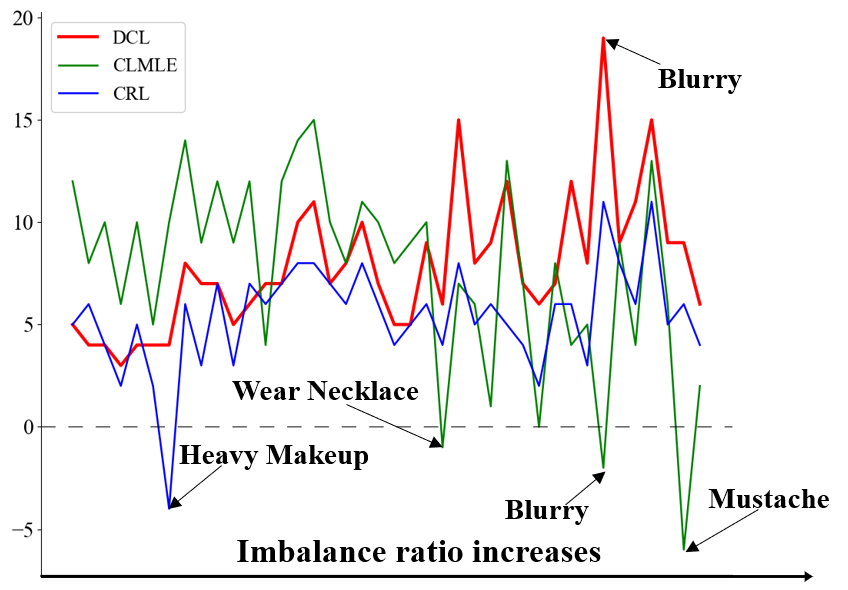}
	\caption{Comparison of performance gain to the DeepID2 for  DCL, CRL and CLMLE with respect to the imbalance ratio.}
	\label{img:ratio}
\end{figure}
%---------------------------------------------------------------------------------

\begin{table}[]
	\centering
	\resizebox{0.45\textwidth}{!}{
		\begin{tabular}{|l|c|c|c|c|}
			\hline
			Method                & SS & TL & LS & Performance \\ \hline
			1: Baseline (DeepID2) & 0  & 0  & 0  & 81.17       \\ \hline
			2: 1 + SS             & 1  & 0  & 0  & 86.58       \\ \hline
			3: 2 + TL             & 1  & 1  & 0  & 87.55       \\ \hline
			4: 3 + LS             & 1  & 1  & 1  & 89.05       \\ \hline
		\end{tabular}
	}
	\vspace{3pt}
	\caption{Ablation study of each component: SS-Sampling Scheduler, TL-Triplet Loss with Easy Anchor, LS-Loss Scheduler.}
	\label{tb:ablation}
\end{table}

\begin{table}[t]
	\centering
	\small
	\begin{tabular}{|l|c|}
		\hline
		Method                                   & Performance    \\ \hline
		1: DeepID2                               & 81.17          \\ \hline
		2: DeepID2 + Convex                      & \textbf{86.58} \\ \hline
		3: DeepID2 + Linear                      & 86.36          \\ \hline
		4: DeepID2 + Concave ($\lambda = 0.99$)  & 85.90          \\ \hline
		5: DeepID2 + Composite                   & 86.07          \\ \hline
		*: DeepID2 + Linear Decreasing Imbalance & 85.11          \\ \hline
	\end{tabular}
	\vspace{3pt}
	\caption{Performance comparison between different scheduler functions selection. Method 2 in this table is corresponding to the method 2 in Table \ref{tb:ablation}.}
	\label{tb:sf}
\end{table}

\begin{table*}[t]
	\centering
	\small
	\begin{tabular}{|c|c|c|c|c|c|c|c|}
		\hline
		Method & Deep-Mar\cite{li2015multi} & Inception-v2\cite{ioffe2015batch} & HP-net\cite{liu2017hydraplus} & JRL\cite{wang2017attribute} & VeSPA\cite{sarfraz2017deep} & LG-Net \cite{liu2018localization} & DCL(ours) \\ \hline
		mA     & 73.8                       & 75.4                              & 76.1                          & 77.8                        & 77.7                        & \2{78.7}                          & \1{83.7}  \\ \hline
	\end{tabular}
	\vspace{3pt}
	\caption{Comaprison with the state-of-the-art methods on RAP\cite{li2016richly} dataset. The 1st/2nd best results are highlighted in red/blue.}
	\label{tb:rap}
\end{table*}

In this part, we show the performance gain of each attribute respecting to the data imbalance level compared with the baseline method DeepID2 in Figure \ref{img:ratio}. In the figure, red, blue, green curves indicate DCL, CRL, CLMLE respectively. The horizontal axis indicates the imbalance level and the vertical axis is the performance gain to the baseline for each method. We can observe that our proposed DCL method stably improves the performance across all the attributes while others degrade in some. Specifically, CRL is poor on attribute `Heavy Makeup'(-4$\%$: level-11) and CLMLE is poor on attributes `Wear Necklace'(-1$\%$: level-43)/`Blurry'(-2$\%$: level-45)/`Mustache'(-6$\%$: level-46). Our method achieves remarkable performance over the other two methods when the data is largely imbalanced, which results from the target distribution transition from imbalanced to balanced in sampling strategy. In the later stage of learning, the model focuses more on minority class while still keeps an appropriate memory for the majority class. The most significantly improved attribute is `Blurry', with imbalance ratio 45 (8$\%$ performance gain to CRL, 21$\%$ to CLMLE). Considering all these three methods adopt the same backbone, results show the advantage of the DCL training framework.

\subsubsection{Ablation Study}
There are several important parts in the proposed DCL framework, including the sampling scheduler, design of the triplet loss with easy anchor and loss scheduler. We provide the ablation study in Table \ref{tb:ablation} to illustrate the advantages of each component. Sampling scheduler (SS) aims to dynamically manage the target data distribution from imbalanced to balanced (easy to hard) and the weight of each sample in $\mathcal{L}_\text{DSL}$ (Equation \ref{eq:dsl}). Triplet loss with easy anchors (TL) modifies the anchor selection of triplet pair for better learning ($\mathcal{L}_\text{TEA}$). Loss scheduler (LS) controls the learning importance between $\mathcal{L}_\text{DSL}$ loss and $\mathcal{L}_\text{TEA}$ loss. From the table, we can see that our two important curriculum schedulers contribute a lot with performance gain to the whole system.

\subsubsection{Effect of Scheduler Function Selection}
Since we design several scheduler functions with different properties, we also include an analysis of them. The experiment setup is that we only include the selection variation for sampling scheduler, disable the metric learning with easy anchor and loss scheduler to avoid the mutual effect. In Table \ref{tb:sf}, remember that the target distribution of methods (2-5) is nonlinearly adjusted by the power operation (Eq. \ref{eq:power}) of the scheduler function value. For method (*), the distribution is simple linearly decreasing to 1 at the end of the training. We can observe that method (*) is much worse than others. Also, the convex function is a better selection for sampling scheduler. According to the definition of scheduler function which indicates the learning speed, it interprets that it is better for the system to  learn the imbalanced data slowly at the very beginning of training and then speed up for balanced data learning.

\subsection{Experiments on RAP Pedestrian Dataset}

\subsubsection{Implementation Details}

\textbf{Network Architecture}
We use ResNet-50\cite{he2016deep} as the backbone for our proposed method. For each attribute, we set an extra feature layer of 64-dimension and a final output layer. Our baseline in table \ref{tb:rap_ratio} is a ResNet-50 model trained with Cross Entropy loss in a multi-task learning framework.

\noindent
\textbf{Hyper-Parameter Settings}
We train DCL with batch size 512, learning rate 0.003, decay at 0.0005 and the epoch at 300. Horizontal Flip is applied during training. Specifically, we set sampling scheduler to convex function in Equation \ref{cos_g}, loss scheduler to composite function in Equation \ref{eq:triplet_weight} with advanced self-learning point $p$  to 0.3, and $k$ in $\mathcal{L}_{TEA}$ (Equation \ref{eq:tea}) to 25.

\subsubsection{Overall Evaluation}
For overall evaluation, we include several the state-of-the-art methods that been evaluated in this dataset, including Deep-Mar \cite{li2015multi}, Inception-v2 \cite{ioffe2015batch}, HP-net \cite{liu2017hydraplus}, JRL \cite{wang2017attribute}, VeSPA \cite{sarfraz2017deep} and LG-Net \cite{liu2018localization}. Table \ref{tb:rap} indicates the average class-balanced mean accuracy (mA) for each method in RAP dataset. The 1st/2nd best results are highlighted in red/blue, respectively. We can see that our proposed DCL method outperforms the previous best one (LG-Net) with a large performance gain (5$\%$).
In term of computational complexity, methods like LG-Net and HP-net apply classwise attention to their model, so their methods take more resource in training and inference. Our proposed method is an end-to-end framework with small extra cost.

\begin{table}[]
	\centering
	\small
	\begin{tabular}{|c|c|c|c|}
		\hline
		Imbalance Ratio (1:x) & 1$\sim$25 & 25$\sim$50 & \textgreater{}50 \\ \hline
		Baseline              & 79.3      & 68.9       & 68.0             \\ \hline
		DCL                   & 83.1      & 83.9       & 85.5             \\ \hline
	\end{tabular}
	\vspace{3pt}
	\caption{Average balanced mean accuracy (mA) in different groups of imbalance ratios. Baseline is a ResNet-50 model trained with cross entropy loss.}
	\label{tb:rap_ratio}

\end{table}

\subsubsection{Effect of Data Imbalance Ratio}
Different from the definition of imbalance level (majority class rate-50$\%$) in CelebA, imbalance ratio (1:x) in RAP is the ratio of minority samples to majority samples. As we mentioned, there are 70 attributes in this dataset and the imbalance ratio is up to 1:1800. Therefore, to show the advantage of our method for imbalanced data learning, we group attributes into three categories concerning  imbalance ratio and compare the average mA with the baseline method. The baseline is a ResNet-50 model trained with cross-entropy loss. From Table \ref{tb:rap_ratio}, we can observe that for group 1 with attribute imbalance ratio from 1$\sim$25, our method outperforms 3.8$\%$ to the baseline. When the data is more imbalance distributed in group 2 with ratio 25$\sim$50 and group 3 with ratio \textgreater{}50, DCL achieves 15.0$\%$ and 17.5$\%$ performance gain, respectively. This result demonstrates that our proposed DCL method indeed works effectively for extremely imbalanced data learning.

\subsection{Experiments on CIFAR-100 Dataset}

To validate the generalization ability of our method, we conduct the experiment on a balanced dataset CIFAR-100 with our learning framework. In this balanced case, methods \cite{drummond2003c4,he2008learning,sl} in Table \ref{tb1} are the same to the baseline method with cross-entropy loss. Also, there is no performance report of LMLE/CLMLE for generalization check. Therefore, we compare the results with the baseline and CRL\cite{crl1} in Table \ref{tb:cifar100}. From the result, we can see our DCL method outperforms the baseline and CRL with $+3.4\%$ and $+2.2\%$, respectively. Compared to CRL, our proposed triplet loss with easy anchor stabilizes the training process. Combined with the loss learning scheduler, DCL makes a better rectification on feature space to provide a better representation for the general classification. 
\begin{table}[]
	\centering
	\resizebox{0.45\textwidth}{!}{
		\begin{tabular}{|c|c|c|c|}
			\hline
			 & Cross Entropy & CRL\cite{crl1}     & DCL(ours)       \\ \hline
			Accuracy  & 68.1     & 69.3 (+1.2) & 71.5(+3.4) \\ \hline
		\end{tabular}
	}
	\vspace{2pt}
	\caption{Results on CIFAR100 dataset (to baseline improvement).}
	\label{tb:cifar100}
\end{table}

\section{Conclusion}
In this work, a unified framework for imbalanced data learning, called Dynamic Curriculum Learning (DCL) is proposed. For the first time, we introduce the idea of curriculum learning into the system by designing two curriculum schedulers for sampling and loss backward propagation. Similar to teachers, these two schedulers dynamically manage the model to learn from imbalance to balance and easy to hard. Also, a metric learning triplet loss with easy anchor is designed for better feature embedding. We evaluate our method on two widely used attribute analysis datasets (CelebA and RAP) and achieve the new state-of-the-art performance, which demonstrates the generalization and discriminative power of our model. Particularly, DCL shows a strong ability for classification when data is largely imbalance-distributed.

{\small
\bibliographystyle{ieee_fullname}
\bibliography{egbib}
}

\end{document}